\documentclass[10pt,twocolumn,letterpaper]{article}

\usepackage{cvpr}              % To produce the CAMERA-READY version
\definecolor{cvprblue}{rgb}{0.21,0.49,0.74}
\usepackage[pagebackref,breaklinks,colorlinks,allcolors=cvprblue]{hyperref}
\usepackage{makecell}
\usepackage{textcomp}
\usepackage{stfloats}
\usepackage{url}
\usepackage{verbatim}
\usepackage{graphicx}
\usepackage{amssymb}
\usepackage{multirow}
\usepackage{tabularx}
\usepackage{wrapfig}
\usepackage{changepage}
\usepackage{setspace}
\usepackage{pifont}
\usepackage{xspace}  % 自动加空格
\usepackage{booktabs}
\usepackage{subcaption}
\usepackage{threeparttable}
\usepackage{xcolor} 
\usepackage{newfloat}
\usepackage{listings}
\usepackage{algorithm}
\usepackage{algorithmic}
\usepackage{amsmath}
\def\confName{CVPR}
\def\confYear{2026}

\title{NoOVD: \underline{No}vel Category Discovery and Embedding for \\ \underline{O}pen-\underline{V}ocabulary Object \underline{D}etection}

\author{
	Yupeng Zhang\textsuperscript{\rm 1,2}\hspace{1em}
	Ruize Han\textsuperscript{\rm 3}\hspace{1em}
	Zhiwei Chen\textsuperscript{\rm 4}\hspace{1em}
	Wei Feng\textsuperscript{\rm 1,2}\hspace{1em}
	Liang Wan\textsuperscript{\rm 1,2}\thanks{Corresponding author.}\hspace{1em}\\
	\textsuperscript{\rm 1}College of Intelligence and Computing, Tianjin University.\\
    \textsuperscript{\rm 2}Key Research Center for Surface Monitoring and Analysis of Relics, State Administration of Cultural Heritage.\\
	\textsuperscript{\rm 3}Faculty of Computer Science and Artificial Intelligence, Shenzhen University of Advanced Technology.\\
	\textsuperscript{\rm 4}School of Artificial Intelligence, Nanchang University.\\
{\tt\small \{zhangyupeng, wfeng, lwan\}@tju.edu.cn, hanruize@suat-sz.edu.cn, zhiweichen@ncu.edu.cn}
}

\begin{document}
\maketitle
\begin{abstract}
Despite the remarkable progress in open-vocabulary object detection (OVD), a significant gap remains between the training and testing phases. 
During training, the RPN and RoI heads often misclassify unlabeled novel-category objects as background, causing some proposals to be prematurely filtered out by the RPN while others are further misclassified by the RoI head. During testing, these proposals again receive low scores and are removed in post-processing, leading to a significant drop in recall and ultimately weakening novel-category detection performance.
To address these issues, we propose a novel training framework—NoOVD—which innovatively integrates a self-distillation mechanism grounded in the knowledge of frozen vision-language models (VLMs).
Specifically, we design K-FPN, which leverages the pretrained knowledge of VLMs to guide the model in discovering novel-category objects and facilitates knowledge distillation—without requiring additional data—thus preventing forced alignment of novel objects with background.
Additionally, we introduce R-RPN, which adjusts the confidence scores of proposals during inference to improve the recall of novel-category objects.
Cross-dataset evaluations on OV-LVIS, OV-COCO, and Objects365 demonstrate that our approach consistently achieves superior performance across multiple metrics.
\end{abstract}

\section{Introduction}
\label{sec:intro}
General object detection techniques have achieved remarkable progress driven by deep neural networks, with methods such as Faster R-CNN~\cite{ren2015faster} delivering outstanding performance. However, their training still heavily depends on extensive manual annotation, requiring tens of thousands of bounding boxes per category—an expensive and inefficient process. Although high-quality datasets like Pascal VOC~\cite{everingham2010pascal} and MS COCO~\cite{lin2014microsoft} are available, the limited number of categories they cover falls far short of human cognitive capabilities. Achieving universal object detection demands exponentially increasing resources, presenting a formidable challenge.

\begin{figure}[t!]
	\centering
	\includegraphics[width=0.9\linewidth]{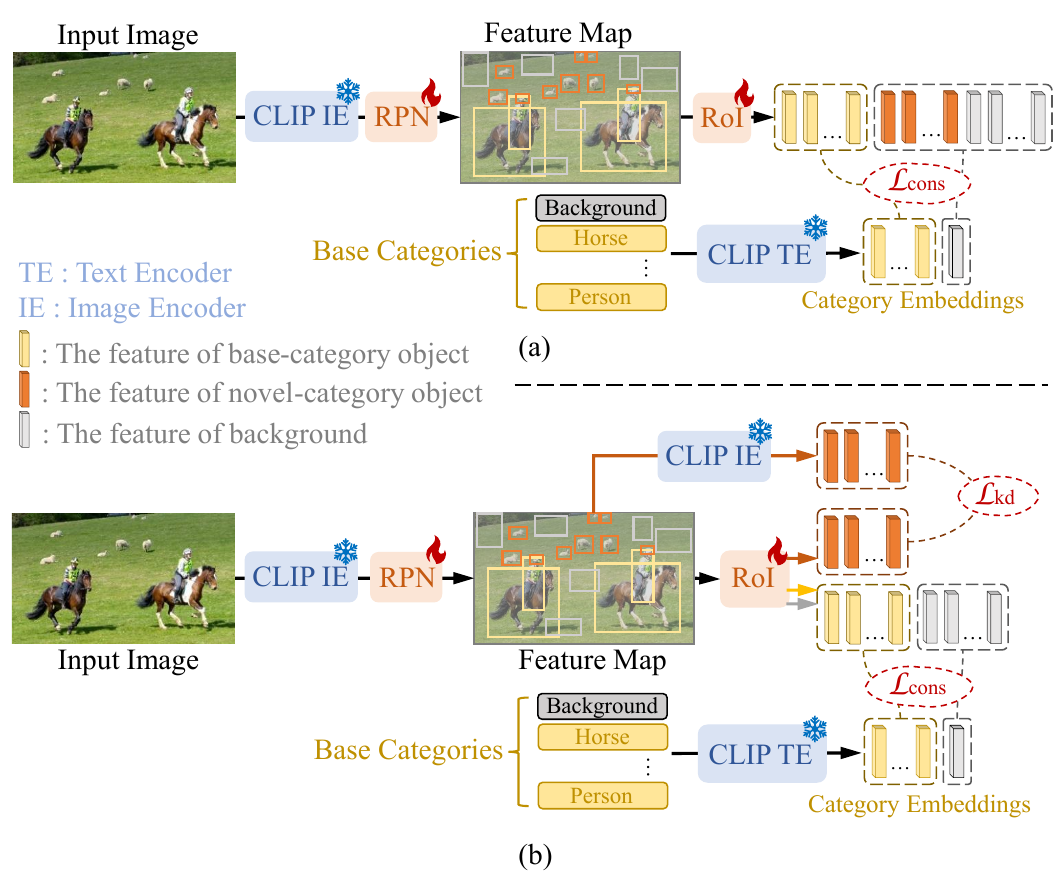}
    \vspace{-10pt}
	\caption{Training process for OVD using frozen CLIP. (a) Commonly used training process, (b) Our training process.} 
	\label{fig:moti}
    \vspace{-20pt}
\end{figure}

In recent years, large-scale vision-language models (VLMs) such as CLIP~\cite{radford2021learning} and ALIGN~\cite{jia2021scaling} demonstrate exceptional zero-shot classification capabilities, driving significant advancements in computer vision. Open-vocabulary object detection (OVD) emerges to overcome the closed-set limitations of traditional detectors. Leading approaches~\cite{gu2021open,pham2024lp,bangalath2022bridging,li2023distilling,du2022learning} rely on VLMs, transferring their zero-shot abilities to detectors through knowledge distillation or image-text embedding alignment. Other methods~\cite{kuo2022f,minderer2022simple,kuo2022f,wang2025ov,wu2023clipself,xu2024dst,wang2025declip} construct detectors based on frozen VLMs, avoiding knowledge degradation during distillation or fine-tuning and preserving generalization to the greatest extent. Although these methods achieve impressive performance in novel category detection, a considerable gap still exists between the training and testing categories.

In the two-stage detection framework built upon a frozen VLM (\eg, CLIP), as shown in Fig.~\ref{fig:moti} (a), the model is trained using only labeled data from base categories. 
During the Region Proposal Network (RPN) training phase, all latent novel-category objects are forcibly regarded as background. 
In the classification phase, the model similarly forces the alignment of novel-category object features with the background text embeddings, which severely hinders the knowledge transfer capabilities of VLMs.
This way, during testing, the proposals for novel-category objects generated by the RPN receive low scores due to being misclassified as background, resulting in their filtration during the RPN's post-processing stage. 
This significantly reduces the detection recall of novel-category objects. 
These limitations ultimately impair detection performance, particularly affecting the recognition of novel-category objects.
Previous methods typically rely on large-scale data training~\cite{kim2023region,wu2023cora,cheng2024yolo,wang2025yoloe} or pseudo-labeling strategies~\cite{zhou2022detecting,bangalath2022bridging,jeong2024proxydet,li2024learning,zhao2024taming,wang2025ov} to discover novel-category objects. However, large-scale data collection and training incur substantial costs, while pseudo-labeling methods depend on text matching and inevitably introduce noise, resulting in limited generalization to novel categories.

As shown in Fig.~\ref{fig:moti}(b), we revisit the OVD training pipeline built on frozen VLMs and introduce an entirely new detection framework.
During training, under supervision from base categories, we first propose the learnable-parameter-free knowledge-retentive FPN (K-FPN), which builds a knowledge-retentive feature pyramid directly from the frozen multi-layer CLIP features, thereby maximizing the retention of CLIP's representation capacity for novel categories. We then project RPN proposals onto K-FPN and, using diverse foreground/background text descriptions generated by an LLM together with CLIP's zero-shot recognition ability, identify latent novel-category proposals and inject their knowledge into the detector through self-distillation—effectively preventing novel-category features from being forced to align with background.
During testing, to prevent novel-category proposals from being prematurely filtered due to low RPN scores, we reuse the training-stage discovery strategy to identify latent novel objects and fuse their foreground scores with the original RPN confidence. The updated proposals are then re-ranked, and the top-K are fed into the RoI head, substantially improving the recall of novel categories.
Notably, unlike previous methods, the entire process requires no additional training data and does not rely on constructing pseudo image–text pairs, thereby eliminating pseudo-label noise.

In summary, the main contributions of this work are:
\begin{itemize}
    \item We propose a novel OVD framework that focuses on the discovery of novel categories. The model simultaneously learns base-category knowledge, identifies latent novel objects, and performs knowledge self-distillation, effectively avoiding the misclassification of novel categories as background and preserving the novel-category knowledge embedded in VLMs.
    
    \item We propose a knowledge-retentive FPN (K-FPN) built on frozen CLIP to preserve world knowledge for novel-category discovery and self-distillation. During testing, we add a Re-weighted RPN (R-RPN) to improve the recall of latent novel-category objects.
    
    \item Comprehensive evaluations on both OV-LVIS~\cite{gupta2019lvis} and OV-COCO~\cite{lin2014microsoft} benchmarks, coupled with cross-dataset validation on Objects365~\cite{shao2019objects365} validation set, consistently demonstrate the SOTA performance, which conclusively establishes the superior robustness and effectiveness of the proposed OVD framework.
\end{itemize}

\section{Related Work}
\label{sec: related}

\begin{figure*}[t!]
	\centering
	\includegraphics[width=0.9\linewidth]{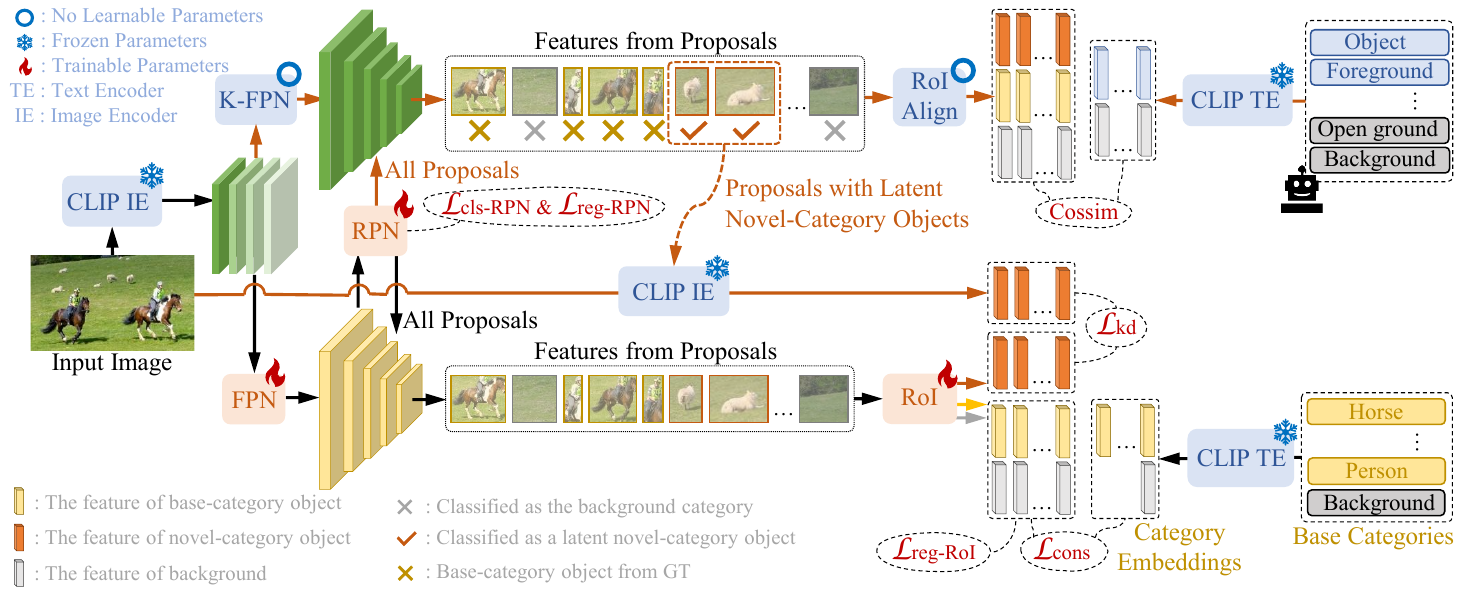}
    \vspace{-7pt}
	\caption{Illustration of the training process of NoOVD. We use K-FPN to extract the pyramid embeddings from frozen CLIP to identify latent novel-category objects. 
    Besides the image-text alignment with $\mathcal{L}_{\text{cons}}$, we also align the features of the RoI head with the features from K-FPN via knowledge self-distillation with $\mathcal{L}_{\text{kd}}$.}
	\label{fig:overall}
    \vspace{-20pt}
\end{figure*}

\textbf{Open-Vocabulary Object Detection (OVD).} With the rapid advancement of vision–language models (VLMs) such as CLIP~\cite{radford2021learning} and ALIGN~\cite{jia2021scaling}, open-vocabulary object detection (OVD) becomes an important research direction, enabling models to recognize both base and novel categories within a unified cross-modal semantic space. Existing approaches largely rely on region–text pairs, knowledge distillation, transfer learning, or pseudo-labeling paradigms: ViLD~\cite{gu2021open} and DetPro~\cite{du2022learning} distill CLIP knowledge into detectors; OADP~\cite{wang2023object} and DK-DETR~\cite{li2023distilling} strengthen semantic modeling during distillation; RO-ViT~\cite{kim2023region}, CORA~\cite{wu2023cora}, YOLO-World~\cite{cheng2024yolo}, and YOLOE~\cite{wang2025yoloe} train on large-scale region–text datasets to achieve efficient open-world perception, \textbf{albeit at high cost.}
Pseudo-labeling methods such as Detic~\cite{zhou2022detecting}, OCO~\cite{bangalath2022bridging}, ProxyDet~\cite{jeong2024proxydet}, LBP~\cite{li2024learning}, SAS-Det~\cite{zhao2024taming}, and OV-DQUO~\cite{wang2025ov} mine potential objects using image-level tags, class-agnostic detectors, or proxy categories; \textbf{however, their reliance on text matching inevitably introduces noise, resulting in limited generalization to novel categories.}
Meanwhile, methods like F-VLM~\cite{kuo2022f}, CLIPSelf~\cite{wu2023clipself}, DST-Det~\cite{xu2024dst}, and DeCLIP~\cite{wang2025declip} build two-stage detectors on frozen CLIP models, training only the detection heads to achieve open-vocabulary recognition. \textbf{Yet, most OVD methods mistakenly treat novel-category objects as background during training and as foreground during inference, leading to a significant training–inference mismatch.}

To address these issues, we leverage the zero-shot capability of frozen VLMs during training to proactively identify latent novel-category objects and introduce a self-distillation mechanism that prevents them from being aligned with background, thereby eliminating noise introduced by pseudo image–text pairs at the source. Notably, our method requires no additional data or changes to the training pipeline while substantially improving novel-category detection performance.

\textbf{Vision-Language Models (VLMs).} Recent years have witnessed groundbreaking advances in VLMs through contrastive learning. Pioneering works such as CLIP~\cite{radford2021learning} and ALIGN~\cite{jia2021scaling} establish unified image-text feature spaces by learning cross-modal alignment on web-scale datasets. These models demonstrate remarkable zero-shot transfer capabilities—for instance, CLIP achieves open-vocabulary classification on ImageNet~\cite{deng2009imagenet} without fine-tuning. This paradigm inspires numerous innovations~\cite{ghiasi2022scaling,rao2022denseclip,xu2023open,zhou2022extract,li2022language,gu2021open,du2022learning,wu2023aligning,wang2023object,kaul2023multi,li2023distilling,kuo2022f,wu2023clipself,zhang2024rethinking,wang2025declip,qian2025covtrack,qiandovtrack,zhang2025odov} for downstream tasks including segmentation and object detection.
We present an OVD framework that efficiently transfers CLIP's pre-trained knowledge. During training, the CLIP encoders are frozen, and their cross-modal alignment is leveraged to uncover latent novel-category objects. A self-distillation mechanism then guides the detector to learn such knowledge, enabling effective transfer while maintaining computational efficiency.

\section{Methodology}
\label{sec:method}
\subsection{Preliminaries and Framework}

Given a target image \( \mathbf{I} \in \mathbb{R}^{3 \times H \times W} \) as input to the detector, two types of outputs are typically desired: 
(1) Location, where the bounding box coordinates \( \mathbf{b}_i \in \mathbb{R}^4 \) represent the location of the \( i \)-th predicted object. 
(2) Category, where \( P_i^j \in \mathcal{C}_{\text{test}} \) representing the \( j \)-th category is assigned to $\mathbf{b}_i$, with \( \mathcal{C}_{\text{test}} \)  as the set of categories during testing.
Following the open-set setting, in the training phase, the categories consist solely of the set of base category \( \mathcal{C}_{\text{base}} \), In the testing phase, the vocabulary is extended to include the set of novel category names \( \mathcal{C}_{\text{novel}} \), $\mathcal{C}_{\text{test}} = \mathcal{C}_{\text{base}} \cup \mathcal{C}_{\text{novel}}$ with \( C_{\text{base}} \cap \mathcal{C}_{\text{novel}} = \emptyset \).

\textbf{Overview.} Our core idea is to fully exploit the strong generalization capabilities of VLMs, \ie. CLIP, which arise from extensive pretraining on large-scale image–text pairs. As shown in Fig.~\ref{fig:overall}, we use frozen CLIP image and text encoders as the backbone and train only the detection modules (FPN, RPN, and RoI) with base-category labels. This structure not only preserves CLIP's pre-trained knowledge but also significantly reduces training costs.

We propose a novel detection framework. During training, we build the detector under supervision from base categories and introduce a learnable-parameter-free architecture, K-FPN, which directly constructs a feature pyramid from frozen multi-layer CLIP features, thereby preserving CLIP's pretrained knowledge to the greatest extent. We then combine diverse foreground–background textual prompts to leverage CLIP's zero-shot capability for discovering latent novel-category objects and integrate this knowledge into the detector through self-distillation.
During testing, we introduce a novel-category retention strategy that identifies latent novel-category objects before RPN post-processing and boosts the confidence scores of their proposals, thereby significantly improving novel-category recall.

%%%%%%%%%%%%%%%%%%%%%%%%%%%%%%%%%%%%%%%%%%%%%%%%%%%%%%%%%%%%%%%%%%%%%%%%%%%%%%%%%%%%%%%%%%%%%%%%%%%%%
\subsection{K-FPN for Knowledge Retention}
\label{sec:K-FPN}
To address the limitation of existing methods that implicitly treat latent novel-category objects as background during training, we propose a new method that simultaneously learns from base-category labels and leverages CLIP's pretrained knowledge.
However, as the model is trained with only limited base-category data, the features extracted by frozen CLIP tend to drift when passing through subsequent modules with learnable parameters (\eg, FPN), making it difficult to fully preserve CLIP's knowledge.
This makes it difficult to detect latent novel-category objects through the generated feature pyramid.
Therefore, a core issue is how to retain the knowledge for novel-category objects discovery.

\begin{figure}[t!]
	\centering
	\includegraphics[width=0.9\linewidth]{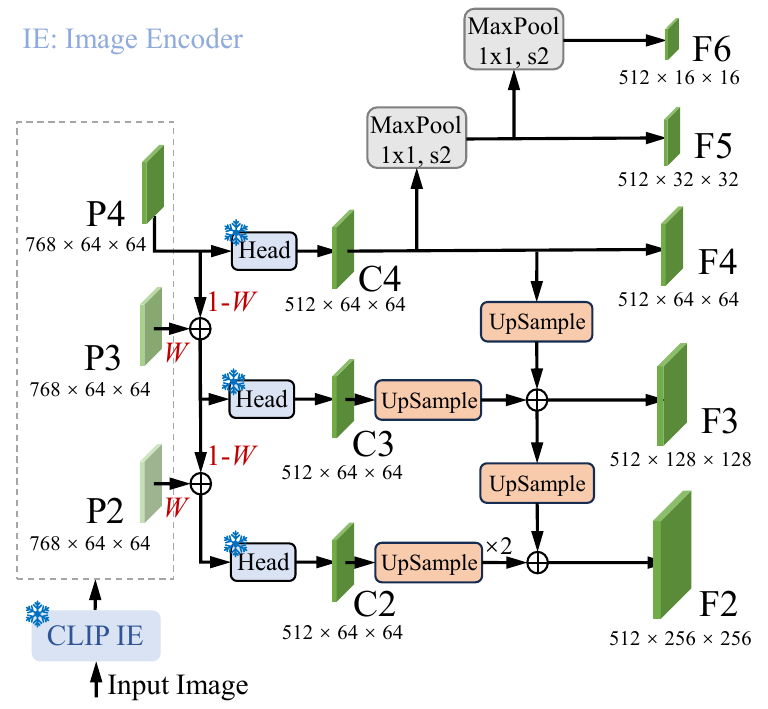}
    \vspace{-10pt}
	\caption{Overall of K-FPN (the CLIP Image Encoder is taken as an example with ViT-B/16).}
	\label{fig:K-FPN}
    \vspace{-15pt}
\end{figure}

To address this, we propose a novel \textit{learnable-parameter-free} network, \textit{\ie}, K-FPN (Knowledge-retentive FPN), that directly utilizes the frozen multi-layer feature maps of CLIP to construct a hierarchical feature pyramid, aiming to preserve CLIP's original knowledge as much as possible. This enables us to effectively discover latent novel-category objects through textual information. The overall architecture is shown in Fig.~\ref{fig:K-FPN}.

Taking CLIPSelf ViT-B/16 as an example, we first extract image features using CLIP's Image encoder, as
\begin{equation}
F_I = \textit{CLIP IE}(\textit{Image}),
\end{equation}
where \textit{CLIP IE} represents CLIP's Image Encoder, \textit{Image} refers to the input image, and $F_I$ represents the feature of the input image.

Following the detection framework F-ViT (proposed in CLIPSelf~\cite{wu2023clipself}), we select the feature maps from layers [5, 7, 11] of $F_i$ as the base feature maps, denoted as \{P2, P3, P4\}, each with a resolution of 768 × 64 × 64. To ensure that each layer retains rich semantic information, we perform top-down feature fusion in an FPN manner. We then employ the \textbf{dimensionality-reduction head} used in the self-distilled CLIP from CLIPSelf, freeze its parameters, and convert the channel dimension of the fused features from 768 to 512 (aligned with the text embedding dimension), yielding the more detail-preserving feature maps \{C2\}, \{C3\}, and \{C4\}.

Then we upsample \{C4\} — containing higher-level semantic information — using bilinear interpolation, a standard feature-processing operation, and subsequently concatenate it laterally with the upsampled \{C3\} and \{C2\} (which, though lower level, provide more precise localization), producing higher-resolution feature maps \{F2, F3, F4\}. Finally, by applying max pooling twice to \{C4\}, consistent with the FPN design, we obtain \{F5, F6\}. This yields a five-level feature pyramid \{F2, F3, F4, F5, F6\}, each layer having 512 channels and sharing the same spatial resolutions as the original FPN.
We represent this process as
\begin{equation}
\mathit{F_{\textit{K-FPN}}} = \textit{K-FPN}(F_I),
\end{equation}
where $\textit{K-FPN}$ represents the process of K-FPN, and $\mathit{F_{\textit{K-FPN}}}$ denotes the feature pyramid generated by K-FPN.
Notably, the entire process involves no learnable parameters, thus maximizing the preservation of CLIP's knowledge, providing a solid foundation for foreground object discovery.

\subsection{Novel Category Discovery and Embedding} 
\label{sec:novel discovery}
Based on the features from K-FPN retaining the knowledge of CLIP, we then consider how to leverage the zero-shot recognition capability of pre-trained large models to uncover latent novel-category objects. It prevents the misalignment of novel-category features with background semantics, thereby mitigating semantic bias. After that, we perform knowledge self-distillation by aligning RoI features with CLIP features, enabling effective knowledge transfer. This process has two main components:

\textbf{General description guided novel category discovery.} 
Given a text prompt, CLIP exhibits impressive zero-shot recognition abilities. However, to effectively identify all foreground objects (both base and novel categories) during the RPN phase, it is necessary to accurately input all category names, which is impractical in real-world applications. Therefore, we design a class-agnostic latent novel-category object discovery strategy, which consists of two steps:

\textit{\ding{172} Foreground-background text descriptions.} To maximize the detection of all foreground objects during the RPN stage, we explore the design of text prompts for foreground objects that are agnostic to specific categories, with the goal of enhancing foreground object detection recall. 
We use ChatGPT-o1 to generate diverse foreground object descriptions as prompts, aiming to detect all foreground objects rather than focusing on any specific category.
We use `object' as the base noun and combine it with higher-level semantic terms (such as `plant' and `animal') to cover a broader range of object categories. The final format is exemplified as: `\texttt{This is an object, specifically a plant}' or `\texttt{This is an object, specifically an animal}'. 
To maintain balance with the number of foreground category descriptions, we also provide various background descriptions, such as `\texttt{This is a background area}' and `\texttt{This is part of the ground}'.
Then, we extract the embedding using the frozen CLIP text encoder,
\begin{equation}
E_T = \textit{CLIP TE} (\textit{Text}),
\end{equation}
where \textit{CLIP TE} represents CLIP's Text Encoder, \textit{Text} refers to the foreground-background text descriptions, and $E_T$ represents the embeddings of the text descriptions.

\textit{\ding{173} Novel-category object discovery.} We directly map the proposals generated by the RPN to K-FPN, crop their features, and extract the proposal features using RoI Align, as
\begin{equation}
F_{\textit{proposals}}^{\textit{K-FPN}} = \textit{RoI Align}\left( \textit{Crop}_{\textit{proposals}}\left( F_{\textit{K-FPN}} \right) \right),
\end{equation}
where \(\textit{Crop}_{\textit{proposals}}\) denotes the cropping operation based on proposals, and \(F_{\textit{proposals}}^{\textit{K-FPN}}\) represents the features obtained by cropping \(F_{\textit{K-FPN}}\) based on proposals.
We then compute the cosine similarity $s$ between these features $F_{\textit{proposals}}^{\textit{K-FPN}}$ and category-agnostic foreground-background text enbeddings $E_T$, as
\begin{equation}
s = \frac{F_{\textit{proposals}}^{\textit{K-FPN}} \cdot E_T}{\|F_{\textit{proposals}}^{\textit{K-FPN}}\| \cdot \|E_T\|}.
\end{equation}
We retain the proposals belonging to the foreground and discard those classified as base categories based on the GT bbox. The remaining proposals are considered to contain latent novel-category objects. 

\textbf{Novel-category object embedding via knowledge self-distillation.}
After selecting these proposals, we crop the corresponding regions from the original image and pass them through the frozen CLIP model to extract features, as
\begin{equation}
F_{\textit{proposals+}}^{\textit{Image}} = \textit{CLIP IE}\left( \textit{Crop}_{\textit{proposals+}}\left( \textit{Image} \right) \right),
\end{equation}
where \(\textit{proposals+}\) denotes the proposals containing latent novel-category objects, and \(F_{\textit{proposals+}}^{\textit{Image}}\) represents the features of the latent novel-category objects extracted by the frozen CLIP Image Encoder.
Subsequently, the features extracted by the frozen CLIP are aligned with the proposal features obtained through RoI, as
\begin{equation}
F_{\textit{proposals+}}^{\textit{RoI}} = \textit{RoI Align}\left( \textit{Crop}_{\textit{proposals+}} \left( F_{\textit{FPN}} \right) \right),
\end{equation}
\begin{equation}
\mathcal{L}_{\text{kd}} = \| F_{\textit{proposals+}}^{\textit{RoI}} - F_{\textit{proposals+}}^{\textit{Image}} \|_2^2,
\end{equation}
where \( F_{\textit{FPN}} \) represents the feature pyramid obtained through FPN, \( F_{\textit{proposals+}}^{\textit{RoI}} \) denotes the features of proposals containing latent novel-category objects after the RoI, and \( \mathcal{L}_{\text{kd}} \) represents the distillation loss.
This process effectively integrates novel-category knowledge into the detector.

% ------------------------------------------------------------------------------
\subsection{Re-Weighted RPN (R-RPN) during Testing}
\label{sec:post}
To mitigate premature filtering of novel-category proposals during RPN post-processing at test time due to low confidence scores, we adjust proposal confidence before post-processing to boost proposals containing latent novel-category objects and improve their recall. Specifically, we first process the NMS results in the RPN. \textbf{We then apply the method in Sec.~\ref{sec:novel discovery} to classify post-NMS proposals as foreground objects}, and combine the resulting confidence with the original RPN confidence as
\begin{equation}
\label{eq:1}
S_{\textit{R-RPN}} = \alpha \cdot S_{\textit{RPN}} + (1 - \alpha) \cdot S_{\textit{K-FPN}},
\end{equation}
where \( \alpha \) is the confidence weight that controls the weighted ratio between the RPN confidence and K-FPN confidence, \( S_{\textit{RPN}} \) represents the foreground object confidence from the original RPN, \( S_{\textit{K-FPN}} \) represents the foreground object confidence from K-FPN, and \( S_{\textit{R-RPN}} \) denotes the confidence of the proposals after the weighted fusion. 

Ultimately, we replace the original RPN confidence with the weighted confidence for subsequent processing steps. 
The subsequent process follows the standard RPN post-processing pipeline: the fused scores are sorted in descending order, and the top 1,000 proposals are retained and passed to the RoI head for classification.
This ensures that the proposals passed to the RoI head effectively encompass latent novel-category objects, thereby enhancing the recall of novel-category objects.

\subsection{Framework and Details}
\textbf{Training stage.} 
To reduce training time and computational overhead, we first adopt the RPN trained on base categories (base-RPN) in CLIPSelf to generate proposals, as base-RPN has been shown in ViLD~\cite{gu2021open} to detect the vast majority of foreground objects. 
Subsequently, following the approach in Sec.~\ref{sec:novel discovery}, we use 30 foreground and 30 background text prompts—an already diverse and balanced set that covers most semantic types in open-world scenarios, with little benefit from further expansion—to select the top 100 candidate boxes whose similarity to foreground prompts exceeds that to background prompts. Choosing 100 candidates strikes an effective balance between coverage and efficiency, capturing potential novel objects while avoiding unnecessary computational and storage overhead.
Next, we rescale these selected proposals to the original image resolution, store them, crop the corresponding image regions, and feed the crops into CLIP to extract and save their features for subsequent knowledge self-distillation. 
During training, if a proposal exhibits high overlap with any cached candidate, we replace it with the cached candidate before feeding it into the RoI head, while the remaining proposals are processed following the standard detection pipeline.

In the self-distillation branch, we use the stored candidates and their features for distillation supervision.
The total loss is
\begin{equation}
\mathcal{L}_{\text{total}} = \mathcal{L}_{\text{reg-RPN}} + \mathcal{L}_{\text{cls-RPN}} + \mathcal{L}_{\text{reg-RoI}} + \mathcal{L}_{\text{cons}} + \mathcal{L}_{\text{kd}}, 
\label{eq:total_loss}
\end{equation}
where $\mathcal{L}_{\text{cls-RPN}}$, $\mathcal{L}_{\text{reg-RPN}}$, and $\mathcal{L}_{\text{reg-RoI}}$ denote the standard classification and regression losses for detection, and $\mathcal{L}_{\text{cons}}$ and $\mathcal{L}_{\text{kd}}$ represent the contrastive loss and knowledge self-distillation loss for the RoI head, respectively. The $\mathcal{L}_{\text{kd}}$ employs the $\mathcal{L}_2$ loss. The weight of the self-distillation loss is set to 1 because we consider novel-category distillation to be as important as base-category alignment, while the remaining loss weights follow the F-ViT.

We adopt the optimized CLIP ViT-B/16 and ViT-L/14 from CLIPSelf~\cite{wu2023clipself} and its improved version DeCLIP~\cite{wang2025declip} as our backbones, and keep them (both the image and text encoders) frozen. During training, we only train the detection modules (FPN, RPN and RoI). Simultaneously, we interpolate the feature maps from layers $[3, 5, 7, 11]$ of ViT-B/16 with relative scales $\left[\frac{1}{4}, \frac{1}{8}, \frac{1}{16}, \frac{1}{32}\right]$ to the input image size. For ViT-L/14, we interpolate the feature maps from layers $[6, 10, 14, 23]$ with relative scales $\left[\frac{1}{3.5}, \frac{1}{7}, \frac{1}{14}, \frac{1}{28}\right]$ to the input image size. In Fig.~\ref{fig:K-FPN}, the weights \( W\) is set to 0.3.
We train the model for 5 epochs on the OV-COCO and for 50 epochs on the OV-LVIS. We use 16 NVIDIA 3090 GPUs, with a batch size of 10 per GPU, and we use the AdamW optimizer with a learning rate of \(10^{-4}\) and a weight decay of 0.1. 

\textbf{Testing stage.}
During inference, we use text prompts for both \textit{base} and \textit{novel} categories. R-RPN recalibrates proposal scores with $\alpha=0.5$ in Eq.~\ref{eq:1}. Object categories are then predicted via cosine similarity between proposal features and all text prompts. Details are provided in the \textbf{supplementary material}.

\section{Experiments}
\label{sec:experiments}

\subsection{Datasets and Evaluation Metrics}
Our method is evaluated on the standard OVD benchmark, LVIS~\cite{gupta2019lvis}. LVIS comprises 100K images and 1,203 categories. The categories are divided into three groups based on the number of training images: `frequent', `common', and `rare'. Following ViLD~\cite{gu2021open}, we treat 337 `rare' categories as \textit{novel categories} and train exclusively on the \textit{base categories} (405 `frequent' and 461 `common' categories). This benchmark is referred to as OV-LVIS. We follow previous work in reporting the average precision for OV-LVIS. we report the average precision for `frequent', `common', and `rare' categories, denoted as $\textit{AP}_\textit{f}$, $\textit{AP}_\textit{c}$, and $\textit{AP}_\textit{r}$ respectively. The symbol $\textit{AP}$ represents the average precision across all categories. The open-vocabulary COCO (OV-COCO) benchmark, proposed by OVRCNN~\cite{zareian2021open}, divides the 65 categories in MS COCO into 48 base and 17 novel categories. Their accuracies are represented by \( \textit{AP}_{\textit{base}}^\textit{50} \) and \( \textit{AP}_{\textit{novel}}^\textit{50} \), while \( \textit{AP}^\textit{50} \) represents the average precision.

\begin{table}[t!]
    \setlength{\tabcolsep}{3.2pt}
	\centering
	% \scriptsize
    % \footnotesize
    \tiny
	\caption{Comparison with SOTA methods on OV-LVIS (\%).} \vspace{-10pt}
    \label{tab:LVIS}
    \begin{threeparttable}
	    \begin{tabular}{l|c|c|c|c|c|c}
        \toprule
        Method    & Backbone & Training Data  &$\textit{AP}_\textit{f}$     &$\textit{AP}_\textit{c}$    &$\textit{AP}_\textit{r}$  &$\textit{AP}$  \\\hline
        \multirow{2}*{RegionCLIP} & {RN50}$^*$    & \multirow{2}*{CC3M}  & 34.0 & 27.4 & 17.1 &  28.2 \\
         & {RN50x4}$^*$ &     & 36.9 & 32.1 & 22.0 & 32.3 \\
        \hline
        Detic  & {RN50}$^*$  & LVIS-base + IN-L & -  & -  &24.9  &32.4\\
        \hline
        \multirow{2}*{OWL-ViT} & ViT-B/16  & \multirow{2}*{O365 + VG}  & - & - & 20.6 &  27.2 \\
         & ViT-L/14   &  & - & - & 31.2 & 34.6 \\
        \hline
        \multirow{4}*{RKDWTF} & {RN50}$^*$ Base   & \multirow{4}*{LVIS-base + IN-L}  & 26.4 & 19.4 & 12.2 & 20.9 \\
         & {RN50}$^*$ RKDPIS   &  & 25.5 & 20.9 & 17.3 & 22.1 \\
         & {RN50}$^*$ WTF  &  & 26.7 & 21.4 & 17.1 & 22.8 \\
         & {RN50}$^*$ WTF8x   &  & 29.1 & 25.0 & 21.1 & 25.9 \\
        \hline
        OADP & ViT-B/32  & LVIS-all  & 32.0 & 28.4 & 21.9 & 28.7 \\ 
        \hline
        DK-DETR & RN50  & LVIS-all  & 40.2 & 32.0 & 22.2 & 33.5 \\
        \hline
        CORA & \multirow{2}*{RN50x4} &LVIS-base  &-  &-  &22.2 &-  \\
        \cline{1-1} \cline{3-3}
        CORA+ &  &LVIS-base + IN-21K  &-  &-  &28.1 &-  \\
        \hline
        \multirow{3}*{RO-ViT} & ViT-B/16 &\multirow{3}*{ALIGN}  &-  &-  & 28.0 & 30.2  \\
        & ViT-L/16 &  &-  &-  & 32.1 & 34.0  \\
        & ViT-H/16 &  &-  &-  & 34.1 & 35.1  \\
        \hline
        \multirow{3}*{YOLO-World} & {YOLOv8-S}$^*$   & \multirow{6}*{O365 + GoldG}  &26.3 &16.3 &13.5 &19.7  \\
         & {YOLOv8-M}$^*$   &  &32.7 &22.5 &19.9 &26.0\\
         & {YOLOv8-L}$^*$   &   &35.4 &24.9 &22.9 &28.7 \\
        \cline{1-2}\cline{4-7}

        \multirow{3}*{YOLOE} & {YOLOv11-S}$^*$   &  &29.3 &26.8 &21.4 &27.5  \\
        & {YOLOv11-M}$^*$   &  &34.5 &32.5 &26.9 &33.0\\
        & {YOLOv11-L}$^*$   &  &36.5 &35.0 &29.1 &35.2 \\
        \hline
        \multirow{6}*{MM-OVOD} & \multirow{6}*{{RN50}$^*$}   & \multirow{3}*{\makecell{LVIS-base}} & - & - &19.3  &30.3  \\
         &   &  & - & - &18.3  &29.2  \\
         &  &  & - & - & 19.3 & 30.6 \\
        \cline{3-7}
         &    &  \multirow{3}*{\makecell{LVIS-base + IN-L}}  & - & - &25.8  &32.7  \\
         &    &    & - & - &23.8  &31.3  \\
         &    &   & - & - & 27.3 & 33.1 \\
                 \hline
           \multirow{4}*{F-VLM} & RN50   & \multirow{22}*{LVIS-base} & - & - & 18.6 & 24.2 \\
         & RN50x4   &  & - & - & 26.3  & 28.5 \\
         & RN50x16   &  & - & - & 30.4 & 32.1 \\
         & RN50x64   &  & - & - & 32.8 & 34.9 \\
         \cline{1-2}\cline{4-7}
        DST-Det & ViT-B/16 &  & -  & -  & 26.2  & -  \\
        \cline{1-2}\cline{4-7}
        \multirow{2}*{SAS-Det} & RN50-C4 &  &31.6  & 26.1 & 20.9  &27.4 \\
	       & RN50x4-C4   &  &36.8  &32.4  &29.1  &33.5 \\
        \cline{1-2}\cline{4-7}
        LBP & - &  &32.4  &28.8   &22.2 &29.1  \\
        \cline{1-2}\cline{4-7}
        \multirow{2}*{OV-DQUO} & ViT-B/16 &  &23.8  &27.7   &29.4  &26.5  \\
        & ViT-L/14 &  &28.5  &36.0  &39.5  &33.7  \\ 

        \cline{1-2}\cline{4-7}
        \multirow{2}*{CLIPSelf + F-ViT} & ViT-B/16  &  &29.1     &21.8 &25.3 &25.2  \\
	       & ViT-L/14   &  &35.6  &34.6  &34.9  &35.1  \\
        \cline{1-2}\cline{4-7}
        \multirow{2}*{CLIPSelf$^\dagger$ + F-ViT} & ViT-B/16  &  &29.3  &21.8 &\underline{25.4} &25.4 \\
	       & ViT-L/14   &  &35.7  &34.8  &\underline{35.0}  &35.2 \\
        \cline{1-2}\cline{4-7}
        \multirow{2}*{CLIPSelf + \textbf{NoOVD (Ours)}}  & ViT-B/16 &  & 30.7  & 22.6  & 28.3 (+2.9)  & 26.7    \\
        & ViT-L/14   &  &37.2   &35.9   & 37.8 (+2.8)  & 36.7  \\

        \cline{1-2}\cline{4-7}
        \multirow{2}*{DeCLIP + F-ViT} & ViT-B/16  &   &29.8   &22.4   &26.8   &26.0  \\
	       & ViT-L/14   &  &36.5   &35.2   &37.2   &36.0  \\
        \cline{1-2}\cline{4-7}
        \multirow{2}*{DeCLIP$^\dagger$ + F-ViT} & ViT-B/16  &  &29.6  &22.5   &\underline{26.6}   &26.0  \\
	       & ViT-L/14   &  & 36.8  & 35.3   & \underline{36.9}  &36.2  \\
        \cline{1-2}\cline{4-7}
        \multirow{2}*{DeCLIP + \textbf{NoOVD (Ours)}} 
        & ViT-B/16 &  &31.2  &23.7   &29.2 (+2.6)   & 27.6   \\
        & ViT-L/14   &  &38.0  &36.9   &39.2 (+2.3)   &37.7  \\
        \bottomrule    
    	\end{tabular}  
        % \vspace{-3pt}
        \begin{tablenotes}
        % \footnotesize
        % \scriptsize
        \tiny
        \item Notes: IN-L denotes the inclusion of images corresponding to the 997 categories shared between ImageNet-21k-P~\cite{ridnik2021imagenet} and LVIS, `$^*$' indicates that the backbone is not initialized with CLIP, and 
        `$^\dagger$' represents the results of our reproduction of CLIPSelf and DeCLIP. 
        CC3M~\cite{sharma2018conceptual}, GoldG~\cite{kamath2021mdetr}, VG~\cite{krishna2017visual}, and ALIGN~\cite{jia2021scaling} are all publicly available datasets.  
        \end{tablenotes}
    \vspace{-3pt}
\end{threeparttable} 
\vspace{-15pt}
\end{table}

\subsection{Comparison Methods} 

For a fair comparison, we select several mainstream VLM-based OVD methods for testing, evaluation, and comparison on the OVD benchmarks. 
Specifically, we include the transfer learning approaches, \ie, OWL-ViT~\cite{minderer2022simple}, F-VLM~\cite{kuo2022f}, CLIPSelf~\cite{wu2023clipself}, DST-Det~\cite{xu2024dst}, LBP~\cite{li2024learning}, OV-DQUO~\cite{wang2025ov} and DeCLIP~\cite{wang2025declip}, and several knowledge distillation methods, \ie, OADP~\cite{wang2023object}, RKDWTF~\cite{bangalath2022bridging}, DK-DETR~\cite{li2023distilling}, RegionCLIP~\cite{zhong2022regionclip}, pseudo-labeling method MM-OVOD~\cite{kaul2023multi}, Detic~\cite{zhou2022detecting}, SAS-Det~\cite{zhao2024taming}, and region-aware training method RO-ViT~\cite{kim2023region}, CORA~\cite{wu2023cora}, YOLO-World~\cite{cheng2024yolo} and YOLOE~\cite{wang2025yoloe}, which retrain a network from scratch using large-scale datasets.
We adopt the optimized CLIP ViT-B/16 and ViT-L/14 from CLIPSelf~\cite{wu2023clipself} and its improved version DeCLIP~\cite{wang2025declip} as our backbones and apply our approach on top of them, using the F-ViT proposed in CLIPSelf as the baseline detection framework.

\subsection{Results on OV-LVIS}
Table~\ref{tab:LVIS} shows the results of all comparative methods and ours on OV-LVIS.

We first observe that NoOVD + DeCLIP ViT-L/14 achieves the best performance among all competitors. Specifically, with ViT-L/14 (304.43M), NoOVD surpasses F-ViT built on the same backbones (CLIPSelf and DeCLIP) by 2.8\% and 1.5\% on `rare' and overall categories, and by 2.3\% and 1.5\% respectively. For the smaller network ViT-B/16 (86.26M), NoOVD also outperforms F-ViT using the same backbones, improving `rare' and overall categories by 2.9\% and 1.3\%, and by 2.6\% and 1.6\% respectively.
These results validate the significant advantage of using proposals containing latent novel-category objects for knowledge self-distillation during training, particularly in enhancing the generalization of novel categories in OVD. 
Notably, NoOVD also shows improvements in base category accuracy compared to F-ViT. We attribute this to the fact that learning novel category knowledge simultaneously strengthens the model’s ability to differentiate between categories. 
Moreover, although methods such as RO-ViT leverage larger-scale pretraining and achieve strong performance even with smaller backbones, NoOVD delivers superior results on novel categories and larger backbones—which are the primary focus of OVD.

\begin{table}[t!] \vspace{-0pt}
    \setlength{\tabcolsep}{3.2pt}
	\centering
	% \scriptsize
    % \footnotesize
    \tiny
	\caption{Comparison with SOTA methods on OV-COCO (\%).} \vspace{-10pt}
    \label{tab:COCO}
	\begin{tabular}{l|c|c|c|c|c}
        \toprule
        Method    & Backbone &Training Data &$\textit{AP}_{\textit{base}}^\textit{50}$     &$\textit{AP}_{\textit{novel}}^\textit{50}$     &$\textit{AP}^\textit{50}$  \\\hline
        \multirow{2}*{RegionCLIP} & RN50$^*$ &\multirow{2}*{CC3M} &57.1  &31.4  &50.4  \\
         &RN50x4$^*$ & &61.6  &39.3  &55.7    \\
        \hline
        Detic   & RN50$^*$ &LVIS-base + IN-L  & 47.1   & 27.8  & 45.0  \\
        \hline
        OWL-ViT & ViT-B/16 &O365 + VG  &-  &-  & 49.2   \\
         % & ViT-L/14 & &-  &-  & 64.7   \\
        \hline
        \multirow{4}*{RKDWTF} & RN50$^*$ Base &\multirow{4}*{LVIS-base + IN-L}  & 53.2 &1.7  &39.6  \\
         & RN50$^*$ RKDPIS & &52.8  &31.5  & 47.2  \\
         & RN50$^*$ WTF &  & 54.0 & 36.6 & 49.4   \\
         & RN50$^*$ WTF8x & &56.6  &36.9  & 51.5  \\
        \hline
        F-VLM & RN50  &LVIS-base  &-  &28.0  &39.6    \\
        \hline
        OADP & ViT-B/32  & LVIS-all  & 53.3 & 30.0 & 47.2 \\ 
        \hline
        DK-DETR & RN50 & LVIS-all &61.1  &32.3  &-   \\
        \hline 
        \multirow{2}*{CORA}  &RN50 &\multirow{2}*{LVIS-base} &35.5 &35.1 &35.4  \\
          &RN50x4 &  &44.5 &41.7 &43.8  \\
        \cline{1-1}\cline{3-3}
        CORA+  &RN50x4 & LVIS-base + COCO Cap. &60.9 &43.1 &56.2  \\
        \hline 
        \multirow{2}*{RO-ViT} &ViT-B/16 & \multirow{2}*{ALIGN} &- &30.2 &41.5  \\
        &ViT-L/16 & &- &33.0 &47.7  \\
        \hline 
        \multirow{2}*{DST-Det} & ViT-B/16 & \multirow{18}*{LVIS-base} &59.6  &41.3  &54.8  \\
        &ViT-L/14 &   &61.9  &46.7  &58.0   \\
        \cline{1-2}\cline{4-6}
        SAS-Det & RN50-C4 &  &58.5  &37.4  &53.0  \\
        \cline{1-2}\cline{4-6}
        LBP & - &  &60.8  &35.9   &54.3  \\
        \cline{1-2}\cline{4-6}
        \multirow{2}*{OV-DQUO} & ViT-B/16 &  &42.6  &39.4   & 41.7  \\
        & ViT-L/14 &  &48.3  &45.3  &47.5  \\ 
        \cline{1-2}\cline{4-6}	
        \multirow{2}*{CLIPSelf + F-ViT} & ViT-B/16 &  &54.9  &37.6  &50.4  \\
	       & ViT-L/14 & &64.1  &44.3  &59.0    \\
        \cline{1-2}\cline{4-6}
        \multirow{2}*{CLIPSelf + F-ViT$^\dagger$} & ViT-B/16 &  &55.8  &\underline{35.8}  &50.6  \\
        &ViT-L/14 &   &64.8  &\underline{42.9}  &59.1   \\
        \cline{1-2}\cline{4-6}

        \multirow{2}*{CLIPSelf + \textbf{NoOVD (Ours)}} & ViT-B/16 & &56.5  &37.9 (+2.1)  &51.6   \\
         & ViT-L/14 & &64.7  &45.4 (+2.5)  &59.7   \\
          % & ViT-L/14 & &63.8  &46.9  &59.4    \\
        \cline{1-2}\cline{4-6}
        
        \multirow{2}*{DeCLIP + F-ViT} & ViT-B/16 &  &57.8  &41.1  &53.5  \\
	       & ViT-L/14 & &65.2  &46.2  &60.3    \\
        \cline{1-2}\cline{4-6}

        \multirow{2}*{DeCLIP + F-ViT$^\dagger$} & ViT-B/16 &  &58.1  &\underline{40.3}  &53.4  \\
	       & ViT-L/14 &  &65.5  &\underline{45.1}  & 60.2   \\
        \cline{1-2}\cline{4-6}
          
        \multirow{2}*{DeCLIP + \textbf{NoOVD (Ours)}} & ViT-B/16 &  &58.3  &41.9 (+1.6)  &54.0   \\
         & ViT-L/14 &  &65.8  &47.5 (+2.4)  & 61.0  \\
	\bottomrule    
	\end{tabular} 
    \vspace{-20pt}
\end{table}
%%%%%%%%%%%%%%%%%%%%%%%%%%%%%%%%%%%%%%%%%%%%%%%%%%%%%%%%%%%%%%%%%%%%%%%%%%%%%%%%%%%%%%%%%%

\subsection{Results on OV-COCO}
Table~\ref{tab:COCO} presents the results of all comparative methods and ours on OV-COCO. 
We first observe that the proposed NoOVD + DeCLIP ViT-L/14 likewise achieves the best performance among all competitors. 
Specifically, when using ViT-L/14 with CLIPSelf and DeCLIP, NoOVD surpasses F-ViT by -0.1\%, 2.5\%, and 0.6\% on the base, novel, and overall categories, and by 0.3\%, 2.4\%, and 0.8\%, respectively. Furthermore, compared to F-ViT built on the CLIPSelf and DeCLIP ViT-B/16 backbones, NoOVD improves the same three groups by 0.7\%, 2.1\%, and 1.0\%, and by 0.2\%, 1.6\%, and 0.6\%, respectively.
We also observe that the performance gains of NoOVD on OV-COCO are smaller than those on OV-LVIS. It is important to note that OV-LVIS is constructed by extending the annotations of OV-COCO, providing richer and more comprehensive labels and covering broader and larger category sets. During training, unlabeled objects in OV-COCO are not treated as background, instead, the model regards them as foreground and distills knowledge from them. However, during testing, although the model successfully detects these objects, they are counted as false positives due to missing annotations, thereby suppressing the overall accuracy. Hence, the smaller performance gains on OV-COCO primarily stem from its incomplete annotations rather than limitations of NoOVD. Therefore, for evaluating OVD, we consider OV-LVIS to be more stable and reliable than OV-COCO.

%%%%%%%%%%%%%%%%%%%%%%%%%%%%%%%%%%%%%%%%%%%%%%%%%%%%%%%%%%%%%%%%%%%%%%%%%%%%%%%%%%%%%%%%%%%

\begin{table}[t!] 
	\centering
	\tiny
	\caption{Cross-dataset main results on Objects365 (\%).} \vspace{-10pt}
    \label{tab:O365}
	\begin{tabular}{l|c|c|c|c|c}
        \toprule
        Method  &Backbone     & Training Data   &$\textit{AP}_\textit{r}$    &$\textit{AP}$   &$\textit{AP}^\textit{50}$\\\hline
        \multirow{2}*{Detic}   & \multirow{4}*{{RN50}$^*$}         & LVIS-all    & 9.5   & 13.9  &19.7  \\
           &      & LVIS-all + IN-L   &12.4    & 15.6  &22.2  \\
        \cline{1-1}\cline{3-6}
        \multirow{2}*{MM-OVOD} &     & LVIS-all   & 10.1 & 14.8  &21.0    \\
         &    & LVIS-all + IN-L   & 13.1 & 16.6  & 23.1   \\
        \hline
	   \multirow{2}*{CLIPSelf + F-ViT} & ViT-B/16    & \multirow{8}*{LVIS-base}   &16.8  & 19.0  &32.3    \\
	   & ViT-L/14   &   &21.7  &23.7  &39.2  \\
        \cline{1-2}\cline{4-6}
        
        \multirow{2}*{CLIPSelf + \textbf{NoOVD (Ours)}} & ViT-B/16   &  &18.3  &19.6  &33.0    \\
         & ViT-L/14   &  &22.8  &24.6  &40.2      \\
        \cline{1-2}\cline{4-6}

        \multirow{2}*{DeCLIP + F-ViT} & ViT-B/16  &   & 17.6  & 20.2   &  33.1  \\
	       & ViT-L/14   &   & 22.3  & 24.5  & 39.8  \\
        \cline{1-2}\cline{4-6}
        
         \multirow{2}*{DeCLIP + \textbf{NoOVD (Ours)}} & ViT-B/16  &  & 19.0 & 21.1 & 34.0    \\
         & ViT-L/14  &   & 23.3  & 25.3   & 41.1       \\
	\bottomrule    
	\end{tabular}
    \vspace{-20pt}
\end{table}

\subsection{Cross-Dataset Transfer Results}

Table~\ref{tab:O365} presents the cross-dataset transfer results from OV-LVIS to Objects365. We compare NoOVD with models from Detic, MM-OVOD, and CLIPSelf / DeCLIP + F-ViT, reporting the bounding box AP metric as the standard on Objects365. In all cases, the Detic and MM-OVOD models are trained on LVIS-all, with IN-L models using ImageNet-21k-P as additional weak supervision, while CLIPSelf / DeCLIP + F-ViT and NoOVD are tested with models trained on the base categories of OV-LVIS. The trained open-vocabulary detectors are evaluated on the Objects365 validation set. Following MM-OVOD, we define the bottom third of the categories in Objects365, based on frequency, as `rare' categories.

Across all settings, CLIPSelf / DeCLIP + F-ViT already surpasses MM-OVOD and Detic. When using ViT-B/16 as the backbone, NoOVD improves over CLIPSelf + F-ViT by 1.5\% in $\textit{AP}_\textit{r}$ and 0.7\% in $\textit{AP}^\textit{50}$, and over DeCLIP + F-ViT by 1.4\% and 0.9\%, respectively. With ViT-L/14, NoOVD outperforms CLIPSelf + F-ViT by 1.1\% in $\textit{AP}_\textit{r}$ and 1.0\% in $\textit{AP}^\textit{50}$, and surpasses DeCLIP + F-ViT by 1.0\% and 1.3\%, respectively.
These results demonstrate that NoOVD, by identifying latent novel-category objects and training the model to adapt to a broader set of categories, significantly enhances the model's transferability.

\subsection{Ablation Study}
Ablation studies use CLIPSelf ViT-B/16 as the backbone.

\textbf{Component analysis.} As shown in Table~\ref{tab:ablation}, We conduct comprehensive ablation experiments on K-FPN and R-RPN using F-ViT as the baseline framework. In the experiments, `CLIP-top' refers to directly using the top-level feature map output by the frozen CLIP (\{P4\}) in Fig.~\ref{fig:K-FPN} to replace K-FPN for feature extraction. 
The results show that using the simple CLIP top-layer feature can improve the performance, which verifies the basic idea of NoOVD.
K-FPN improves novel category detection more significantly by 2.1\%, demonstrating that K-FPN, by constructing multi-scale feature maps, is more effective in discovering latent novel-category objects and facilitating knowledge self-distillation in the network. 
Meanwhile, the use of R-RPN directly enhances the baseline method by 1.3\% in novel category detection, indicating that improving the recall rate of novel-category objects in the RPN stage during testing is essential. This helps prevent novel-category objects from being filtered out during detection, which would otherwise affect subsequent classification. Ultimately, the combination of K-FPN and R-RPN yields the best performance. 

\begin{table}[t!]
	\centering
    % \footnotesize
    \scriptsize
	\caption{Effects of each module in NoOVD on the OV-LVIS (\%).}\vspace{-10pt}
    \label{tab:ablation}
	\scalebox{1.0}{
		\begin{tabular}{cccc|c|c}
			\toprule
			Baseline    &CLIP-top    & K-FPN        & R-RPN           & $\textit{AP}_\textit{r}$       & $\textit{AP}$  \\\hline
			\checkmark  &             &              &                 & 25.4         & 25.4   \\
			\checkmark  & \checkmark  &              &                 & 26.4         & 26.1   \\
            \checkmark  & \checkmark  & \checkmark   &                 & 27.5         & 26.4   \\
			\checkmark  &             &              & \checkmark      & 26.7         & 25.9   \\
            \checkmark  & \checkmark  &              & \checkmark      & 27.8         & 26.4   \\
			\checkmark  & \checkmark  & \checkmark   & \checkmark      & 28.3         & 26.7   \\
			\bottomrule
	\end{tabular}}\vspace{-20pt}
\end{table}

\begin{table}[h!] \vspace{-10pt}
    \centering
    \scriptsize
    \caption{Sentitivity analysis of hyperparameter $W$ (\%).}\vspace{-10pt}
    \label{tab:w}
    \begin{tabular}{c|cccccc}
    \toprule
    \textbf{$W$} & 0.2 & 0.3  & 0.4 & 0.5 & 0.7 & 0.8 \\
    \midrule
    \textbf{$\textit{AP}$} & 26.2  & 26.7   & 26.3    & 24.8    & 22.5   & 19.1  \\
    \bottomrule
    \end{tabular}\vspace{-10pt}
\end{table}

\textbf{Feature fusion weights of K-FPN.}
We conduct ablation studies on the fusion weights of different K-FPN feature levels. As shown in Table~\ref{tab:w}, the best performance occurs at $W = 0.3$. 
When $W$ is too small, high-level features dominate excessively and suppress low-level details, leading to performance degradation; when $W$ is too large, high-level semantic information becomes insufficient, and accuracy similarly drops. 
This confirms that high-level features provide strong semantic abstraction, whereas low-level features contribute essential structural and textural details.
This observation is consistent with prior understanding: high-level features offer stronger semantic abstraction, whereas low-level features preserve richer local details, making them naturally complementary. 
An appropriate fusion ratio strikes an effective balance between semantics and detail, significantly improving overall performance.

\textbf{The fusion weight of R-RPN.}
We conduct ablation studies on the fusion weight $\alpha$ in Eq.~\ref{eq:1}. As shown in Table~\ref{tab:a}, $\alpha = 0.5$ yields the best performance, which also aligns with our intuition. We argue that the K-FPN and RPN scores play equally important roles in R-RPN, and assigning them equal weights establishes a balanced and effective fusion mechanism. Under this strategy, base-category objects retain high fused scores because both K-FPN and RPN provide strong confidence, ensuring they are reliably forwarded to the RoI stage. For latent novel-category objects, although their RPN scores are often low, the strong semantic cues from K-FPN—derived from VLM knowledge—elevate their fused scores, substantially increasing their likelihood of being preserved. Meanwhile, proposals classified as background by both branches maintain low fused scores and are naturally filtered out. Overall, $\alpha = 0.5$ strikes an effective balance between base and novel categories, preserving reliable detection of known classes while significantly improving the recall of latent novel objects.
\begin{table}[t!]
    \centering
    \scriptsize
    \caption{Sentitivity analysis of hyperparameter $\alpha$ (\%).}\vspace{-10pt}
    \label{tab:a}
    \begin{tabular}{c|ccccc}
    \toprule
    \textbf{$\alpha$} & 0.9  & 0.7  & 0.5  & 0.3  & 0.1\\
    \midrule
    \textbf{$\textit{AP}$} &25.5   &25.8    & 26.7    & 26.2    &25.9      \\
    \bottomrule
    \end{tabular}\vspace{-20pt}
\end{table}

\textbf{Choice of distillation loss function.}
As shown in Table~\ref{tab:Lkd}, we conduct an ablation study on the loss function $\mathcal{L}_{\text{kd}}$ used for knowledge self-distillation. We explore several loss functions, including $\mathcal{L}_{\text{1}}$, $\mathcal{L}_{\text{2}}$, $\mathrm{Smooth}\mathcal{L}_{\text{1}}$, and $\mathcal{L}_{\text{cons}}$ (cosine similarity loss). The experimental results indicate that $\mathcal{L}_{\text{2}}$ yields the best performance, while $\mathcal{L}_{\text{1}}$ and $\mathrm{Smooth}\mathcal{L}_{\text{1}}$ also produce similar results. However, using $\mathcal{L}_{\text{cons}}$ for alignment performs relatively poorly. 
We argue that the $\mathcal{L}_{\text{2}}$, being more sensitive to large deviations, can more effectively minimize the distance between RoI features and CLIP representations, thereby promoting precise alignment within the semantic space. In contrast, $\mathcal{L}_{\text{1}}$ treats all errors uniformly, while $\mathrm{Smooth}\mathcal{L}_{\text{1}}$ is overly gentle in regions of small discrepancy, potentially hindering the convergence of cross-modal differences.
Moreover, training with $\mathcal{L}_{\text{cons}}$ may fail to sufficiently align the feature representations, leaving a significant information gap between them. This misalignment can impair the acquisition of novel category knowledge and ultimately compromise detection performance.

\begin{table}[h!] \vspace{-8pt}
	\centering
    \scriptsize
	\caption{Results of different knowledge distillation losses (\%).}
	\vspace{-10pt}
	\label{tab:Lkd}
	\begin{tabular}{c|cccc}
		\toprule
		  \textbf{$\mathcal{L}_{\text{kd}}$}   &\textbf{$\mathcal{L}_{\text{1}}$}  &\textbf{$\mathcal{L}_{\text{2}}$}  &\textbf{$\mathrm{Smooth}\mathcal{L}_{\text{1}}$} &\textbf{$\mathcal{L}_{\text{cons}}$}  \\\hline
		$AP_r$ (\%)   &27.8 &28.3 &28.0 &17.3  \\
		\bottomrule 
	\end{tabular}
    \vspace{-12pt}
\end{table}

\textbf{Extra computation cost.} 
NoOVD's extra cost compared to F-ViT comes from proposal cropping and feature extraction, both carried out offline before training. We use an RPN trained on base categories (replaceable by a stronger variant) to generate proposals and extract all proposal features in one pass using CLIP ViT-B/16 on 8×3090 GPUs (53 minutes in total). Because each image's novel-category proposals are fixed, training only loads cached features for distillation, requiring no repeated cropping or forward passes. Other than this offline step, the overall training time is effectively the same as F-ViT.

\textbf{Additional computation cost analysis and visualizations are included in the supplementary materials.}

\section{Conclusion}
\label{sec:conclusion}
In this work, we have thoroughly examined the issues present in current OVD frameworks, particularly the category gap between training and testing phases.
To this end, we propose an novel framework, NoOVD, which leverages the knowledge from frozen VLMs to uncover latent novel-category objects, while integrating knowledge self-distillation to prevent the forced alignment of novel categories with the background. 
This approach allows the model to learn novel-category knowledge based on the latent novel-category objects. 
During testing, we have enhanced the recall rate of novel-category objects by adjusting the confidence of proposals. 
Extensive experiments validate the effectiveness of NoOVD, demonstrating superior performance. 
Our research introduces a new schema for OVD.

\section*{Acknowledgements}
This work was supported in part by the National Natural Science Foundation of China under Grants U2574216, 62402490, and 62506146; in part by the Emerging Frontiers Cultivation Program of Tianjin University Interdisciplinary Center; in part by the Guangdong Basic and Applied Basic Research Foundation under Grant 2025A1515010101; in part by the Jiangxi Provincial Natural Science Foundation under Grant 20252BAC200196.

{
    \small
    \bibliographystyle{ieeenat_fullname}
    \bibliography{main}
}

\end{document}